\documentclass[conference]{IEEEtran}
\IEEEoverridecommandlockouts
\usepackage{cite}
\usepackage{amsmath,amssymb,amsfonts}
\usepackage{algorithmic}
\usepackage{graphicx}
\usepackage{textcomp}
\usepackage{xcolor}

\usepackage{booktabs}
\usepackage{multirow}
\usepackage[numbers,sort&compress]{natbib}
\usepackage{comment}
\usepackage{hyperref}
\usepackage{color}

\usepackage{makecell}
\usepackage{colortbl}
\usepackage[acronym, nopostdot, nonumberlist]{glossaries}
\usepackage{xpatch}

\makenoidxglossaries
\glsdisablehyper 
\newacronym{ssl}{SSL}{Self-Supervised Learning}
\newacronym{sota}{SOTA}{state-of-the-art}
\newacronym{map}{mAP}{mean Average Precision}
\newacronym{iou}{IoU}{Intersection over Union}
\newacronym{simclr}{SimCLR}{Simple Framework
for Contrastive Learning}
\newacronym{dclw}{DCLW}{Weighted Decoupled Contrastive Learning Loss}
\newacronym{moco}{MoCo}{Momentum Contrast}
\newacronym{byol}{BYOL}{Bootstrap Your Own Latent}
\newacronym{fastsiam}{FastSiam}{Fast Siamese}
\newacronym{simsiam}{SimSiam}{Similarity Siamese}
\newacronym{dino}{DINO}{Distillation with No labels}
\newacronym{arcface}{ArcFace}{Additive Angular Margin}
\newacronym{supcon}{SupCon}{Supervised Contrastive}
\newacronym{knn}{KNN}{$k$-nearest neighbour}
\newacronym{miou}{mIoU}{mean Intersection over Union}
\newacronym{pck}{PCK}{Percentage of Correct Keypoints}
\newacronym{vit}{ViT}{Vision Transformer}
\newacronym{ak}{AK}{Animal Kingdom}
\newacronym{dti}{(DTI}{Department of Trade
and Industry}
\newacronym{cair}{CAIR}{Center of Artificial Intelligence Research}

\xpretocmd{\section}{\glsresetall}{}{}

\def\BibTeX{{\rm B\kern-.05em{\sc i\kern-.025em b}\kern-.08em
    T\kern-.1667em\lower.7ex\hbox{E}\kern-.125emX}}
\begin{document}

\title{Wildlife Target Re-Identification Using Self-supervised Learning in Non-Urban Settings
\thanks{Funding supported by the Department of Trade and Industry (DTI) CAIR.}
}

\author{\IEEEauthorblockN{Mufhumudzi Muthivhi}
\IEEEauthorblockA{\textit{Institute for Intelligent Systems} \\
\textit{University of Johannesburg}\\
Johannesburg, South Africa \\
15mmuthivhi@gmail.com}
\and
\IEEEauthorblockN{Terence L. van Zyl}
\IEEEauthorblockA{\textit{CAIR}, \textit{Institute for Intelligent Systems} \\
\textit{University of Johannesburg}\\
Johannesburg, South Africa \\
tvanzyl@uj.ac.za}
}

\maketitle

\begin{abstract}
Wildlife re-identification aims to match individuals of the same species across different observations.
Current \gls{sota} models rely on class labels to train supervised models for individual classification. This dependence on annotated data has driven the curation of numerous large-scale wildlife datasets.
This study investigates self-supervised learning \gls{ssl} for wildlife re-identification. 
We automatically extract two distinct views of an individual using temporal image pairs from camera trap data without supervision. The image pairs train a self-supervised model from a potentially endless stream of video data. We evaluate the learnt representations against supervised features on open-world scenarios and transfer learning in various wildlife downstream tasks.
The analysis of the experimental results shows that self-supervised models are more robust even with limited data. Moreover, self-supervised features outperform supervision across all downstream tasks. The code is available here \href{https://github.com/pxpana/SSLWildlife}{https://github.com/pxpana/}.
\end{abstract}

\begin{IEEEkeywords}
wildlife, re-identification, self-supervised learning, open-world learning, transfer learning
\end{IEEEkeywords}

\section{Introduction}

Wildlife re-identification aims to recognize and match individual animals across diverse conditions. Re-identification has proven valuable for wildlife conservation, behavioural research, and animal population management~\cite{saoud2024beyond}. It relies on extracting fine-grained local features, such as unique markings, patterns, or subtle variations in color~\cite{crall2013hotspotter}.
Deep learning has successfully automatically extracted more robust and generalizable features for wildlife re-identification~\cite{dlamini2020automated, cermak2024wildfusion, jiao2023toward}. The majority of this work uses supervised learning over class labels. Recent research emphasizes the importance of creating adequately annotated datasets. Towards this objective, \v{C}erm{\'a}k \textit{et al.}~\cite{vcermak2024wildlifedatasets} collect and release several wildlife datasets in an open-source toolkit. Later, Luk{\'a}{\v{s}~\textit{et al.} \cite{adam2024wildlifereid} pre-processed the datasets under one unified pipeline to produce a standard benchmark for training and evaluation. In addition, Lasha \textit{et al.}~\cite{otarashvili2024multispecies} introduces a multi-species dataset to help identify animals with limited data. 
\gls{ssl} does not require an annotated dataset for learning representations. Instead, \gls{ssl} generates supervision signals by creating augmented views of the same image~\cite{wu2018unsupervised}. We propose leveraging camera trap data to extract temporal views for self-supervised training automatically.
\begin{figure}[htp]
  \centering
  \includegraphics[width=0.95\columnwidth]{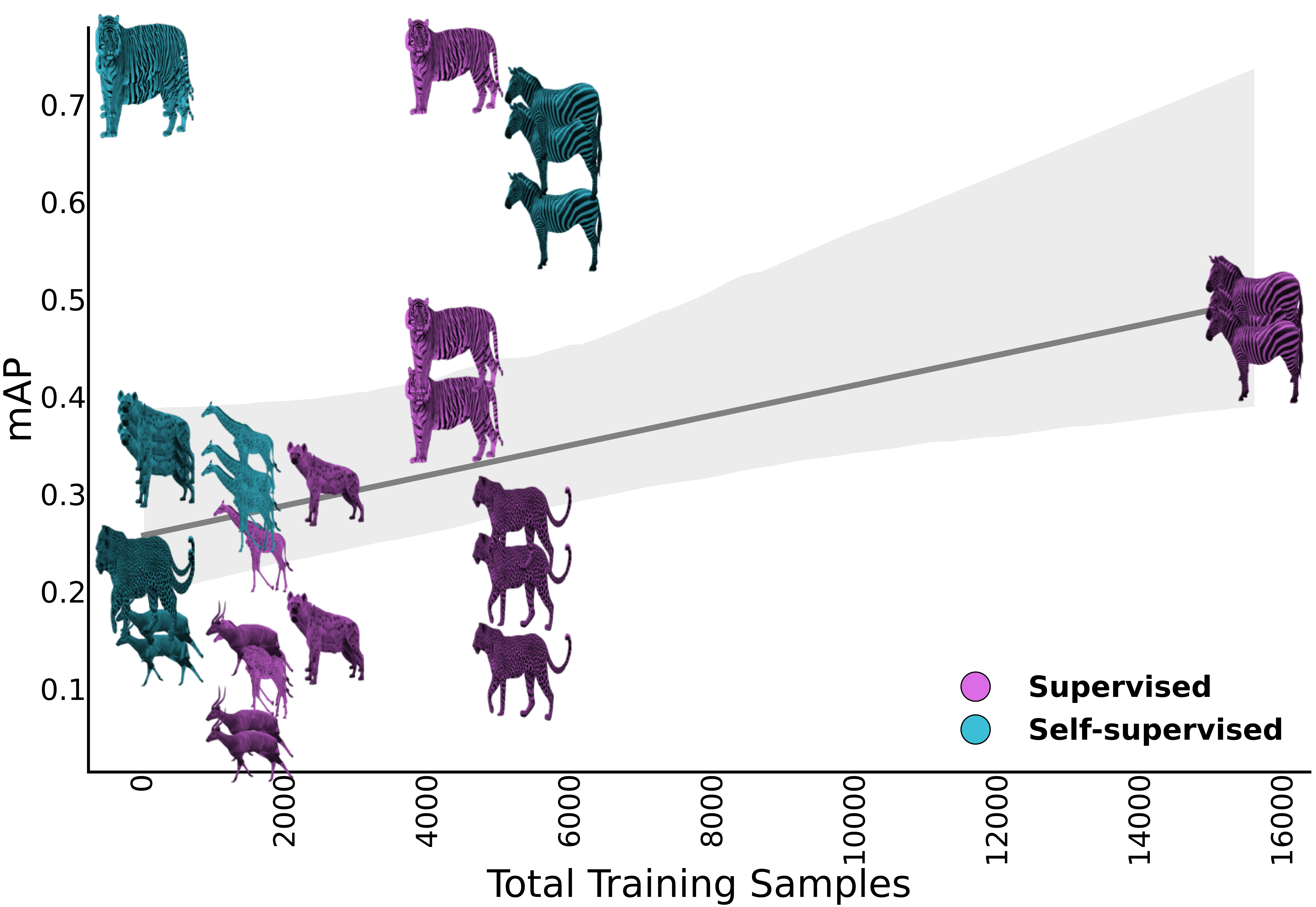} 
  \caption{Comparison of Supervised (Purple) and Self-Supervised (Blue) Models: A scatter plot illustrating the relationship between the total number of training samples and \gls{map} across various models. A robust regression line (grey) highlights the overall trend. Notably, most \gls{ssl} models (blue animal icons) outperform the trend, achieving higher accuracy with fewer samples. For instance, three \gls{ssl} models achieved mAP of 0.5 to 0.7 using 6,000 samples, while supervised models reached 0.35 to 0.5 mAP despite using 16,000 samples.}\label{figure:performance}
\end{figure}
This strategy produces more robust representations and stronger transfer learning ability to downstream tasks.
Figure~\ref{figure:performance} demonstrates the performance of three \gls{ssl} strategies against supervised learning methods. In particular, contrastive learning, self-distillation and the current \gls{ssl} \gls{sota} achieve higher accuracy with exposure to fewer animals than supervised models for open-world data.
Furthermore, results show that self-supervised representations outperform supervised representations for all downstream tasks and out-of-domain data.
We contribute to the existing literature by:
\begin{itemize}
    \item proposing a novel temporal image-pair-based strategy to extract training data for \gls{ssl} automatically;
    \item conduct a large-scale comparison of supervised and self-supervised models over open-world wildlife data, transfer learning and out-of-domain benchmarks;
    \item conduct an ablation study on the effects of training data size on wildlife (\gls{ssl}) models; and
    \item providing a qualitative analysis of supervised and self-supervised features and their attention maps.
\end{itemize}

\section{Related Work}\label{section:RelatedWork}

\subsection{Supervised learning}

Early approaches to wildlife re-identification utilized local feature-based methods~\cite{lowe2004distinctive, detone2018superpoint}. These strategies identify unique key points and extract local descriptors to match individuals against a database. However, these methods do not capture global patterns and often require extensive tuning. Later, practitioners made use of pre-trained general-purpose models. BioClip is a species classification model~\cite{stevens2024bioclip}. It is trained on the TreeOfLife-10M dataset, which encompasses various animals, plants, fungi and insects. 
TransReID is a general-purpose re-identification model primarily used for vehicles and persons~\cite{he2021transreid}. They prioritize generating robust features by shifting and shuffling patch embeddings. 
The current \gls{sota} animal re-identification methods are trained specifically on wildlife datasets. MegaDescriptor is a foundational transformer-based model trained on a large-scale, diverse wildlife dataset~\cite{vcermak2024wildlifedatasets}. Their work was the first to show the representative superiority of wildlife re-identification features over pretrained general-purpose models. MiewID uses contrastive learning with \gls{arcface} loss similar to MegaDescriptor~\cite{otarashvili2024multispecies}. They use a multi-species dataset to pull similar animal individuals closer in the embedding space. 

\subsection{Self-Supervised learning}

Aleksandr~\textit{et al.}~\cite{algasov2024understanding} conducted several experiments on \gls{sota} supervised re-identification models and found that the size of the training data contributes substantially to the performance of deep learning models. However, manually annotating an ever-increasing stream of re-identification data is infeasible~\cite{plum2023replicant}. \gls{ssl} can extract robust features without any annotations~\cite{asano2019critical}.  
Several enhancements to the model architecture and loss function formulation have been proposed for \gls{ssl}. Most of them have reported success over their supervised counterparts.
Contrastive learning methods minimize the distance between positive pairs within the embedding space while simultaneously trying to keep negative pairs apart~\cite{chen2020simple, he2020momentum}. 
Self-distillation methods forgo the need for negatives by using a predictor network to predict the representation of views from the same image. The negative cosine similarity is maximized to pull representations of different views closer together~\cite{chen2021exploring, grill2020bootstrap}. Other methods de-correlate the features of different views by minimizing the redundancy between components~\cite{zbontar2021barlow}. Most of these methods can benefit by conducting multiple crops over a single image to produce global and local views~\cite{caron2021emerging, caron2020unsupervised}. 

\subsection{Representation Learning}

Experiments show that models achieving higher classification accuracy produce representations that can transfer well across various downstream tasks~\cite{kornblith2019similarity}. Supervised learning is known to produce high in-distribution accuracy. However, its performance diminishes when evaluated on multiple out-of-distribution and out-of-domain datasets~\cite{liu2021towards}. Concurrently, self-supervised features have been shown to improve open-world learning~\cite{dhamija2021self}. \gls{ssl} also conveys stronger transferability characteristics compared to supervised learning methods~\cite{ericsson2021well, caron2020unsupervised}. 
Most experiments in wildlife re-identification are conducted over a closed-world setting with supervised learning~\cite{crall2013hotspotter, dlamini2020automated, cermak2024wildfusion, jiao2023toward}. Few have examined open-world learning by testing the model's performance on unseen classes~\cite{vcermak2024wildlifedatasets, adam2024wildlifereid, otarashvili2024multispecies}. To our knowledge, no study has evaluated the open-world performance and transferability of representations over wildlife datasets. This study investigates the robustness and generalization ability of self-supervised and supervised representations for wildlife learning. 

\begin{figure*}[htb!]
\centerline{\includegraphics[width=0.86\textwidth]{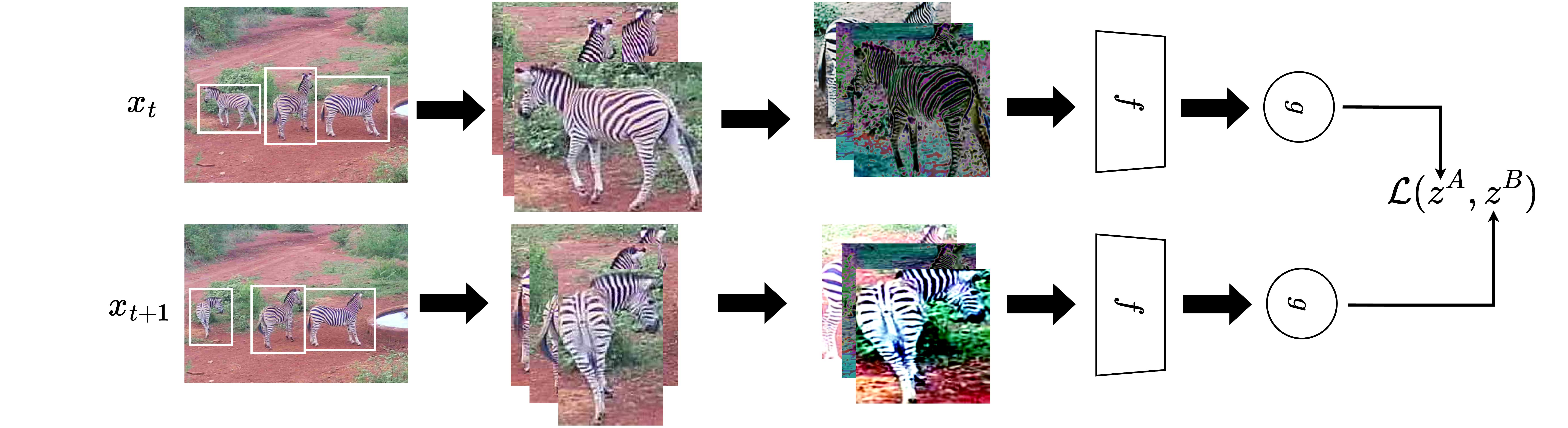}}
\caption{An illustration of extracting two image pairs from consecutive camera frames. The animals are detected with a bounding box around them then we apply IoU to find the same individual in the next frame. Each pair is augmented and passed through an encoder $f$ and projection head $p$ where the loss is computed according to the chosen self-supervised learning (SSL) method.}
\label{figure:model}
\end{figure*}

\section{Methodology}\label{section:Methodology}

\subsection{Self-Supervised Learning}

\gls{ssl} aims to ensure invariance between different views of the same image~\cite{wu2018unsupervised}. Let $x$ be an input image sampled from dataset $\mathcal{D}$. Two random augmentations are applied to obtain $x^{A}$ and $x^{B}$. Given a backbone encoder $f$ and projection head $g$ we extract feature representations:
\begin{equation}
    z = (f \circ g) (x).
\end{equation}
The objective is to minimize the loss over $z^{A}$ and $z^{B}$ such that
\begin{equation} \label{equation:loss}
    \arg\min_{\theta} \mathbb{E}_{x \sim \mathcal{D}} [\mathcal{L} (z^{A}, z^{B})]
\end{equation}
where $\theta$ are the parameters of the backbone and projection head. Several image augmentations, such as random cropping, colour jittering, Gaussian blur grayscaling and horizontal flip, are applied to the input image to simulate basic variations in lighting conditions and viewpoint changes. 

\subsection{Temporal Image Pairs}

Camera trap data often contains several frames of the same animal with different views. Hence, camera traps can automatically generate several augmentations.
Let $\mathcal{T}$ be a sequence of camera trap frames arranged in temporal order. We apply an object detection model, MegaDetector, to locate various animals in $\mathcal{T}$~\cite{beery2019efficient}. Let $x_t$ be a cropped image of an animal at time $t$ such that $x_t \in \mathcal{T}$. Next, we must locate the same individual within the next frame $x_{t+1}$. However, the animal in $x_{t+1}$ has likely moved from its exact position in $x_t$. We use \gls{iou} under some threshold $\alpha$ to find the precise animal within the next frame $x_{t+1}$. Figure~\ref{figure:model} depicts extracting temporal pairs and eventually using them within the loss function in Equation~\ref{equation:loss}. From video frames, we can generate image augmentations such that $x_t \sim x^{A} \text{ and } x_{t+1} \sim x^{B}$ resemble the same individual $x$ but under two separate views.

\subsection{Methods}

We explore six popular self-supervised learning methods and endow them with temporal data.

\textbf{\gls{simclr}} uses positive pairs to enforce invariance to augmentations and samples negatives from the mini-batch to prevent the model from collapsing~\cite{chen2020simple}. Various contrastive loss variants have been proposed for \gls{simclr}. This study selects Weighted \gls{dclw}, which removes the positive term from the denominator and uses a negative von Mises-Fisher weight $w$ to adjust the contribution of each negative pair~\cite{yeh2022decoupled}.
\textbf{\gls{moco}} adopt a queue of negative samples generated by a momentum encoder~\cite{he2020momentum}. They use a stop gradient on the negative samples to stop them from undergoing gradient updates and the NTXent contrastive learning loss function~\cite{sohn2016improved}. \textbf{BarlowTwins} enforces consistency of positive pairs by regularizing the co-variance matrix between two augmented pairs~\cite{zbontar2021barlow}. A cross-correlation loss function is used to decorate the components of the feature space. \textbf{\gls{byol}} has two identical encoders whose weights are not shared~\cite{grill2020bootstrap}. The online encoder has a moving average over the weights to update the target encoder. A negative cosine similarity loss maximizes the similarity between two augmented views passed through two separated encoders and a predictor $h$. \textbf{\gls{fastsiam}} is a faster, more efficient version of the siamese similarity learning-based network \gls{simsiam}~\cite{pototzky2022fastsiam}. They also use a stop gradient operation on the online encoder to prevent the model from collapsing into degenerate solutions. \textbf{\gls{dino}} leverages a student-teacher framework with multi-cropping such that global and local views are generated from the augmentations~\cite{caron2021emerging}. 
A cross-entropy loss function is applied over the predicted features and those generated by the teacher network.

\section{Experimental Setup}\label{section:ExperimentalSetup} 

\begin{figure}[htb!]
\centerline{\includegraphics[trim={0 64.27cm 0 0},clip, width=\columnwidth]{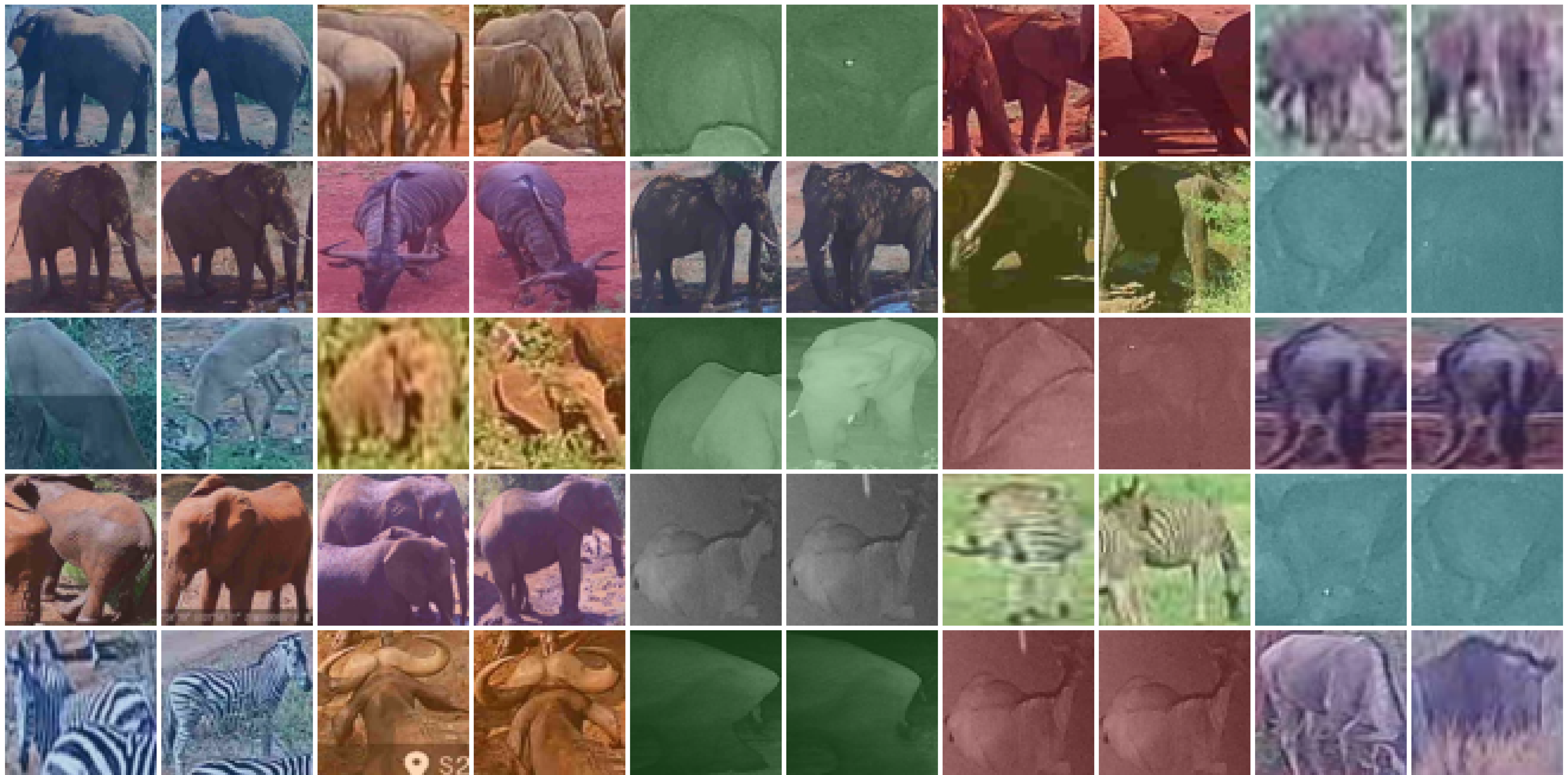}}
\caption{A sample of temporal pairs, where each pair is highlighted with the same colour to indicate two views of the same individual.}
\label{figure:pairs}
\end{figure}

\subsection{Baselines}

The study will use three supervised learning methods. \textbf{\gls{arcface}} loss has produced \gls{sota} results for MegaDescrptor and MiewID~\cite{vcermak2024wildlifedatasets, deng2019arcface, otarashvili2024multispecies}. 
It enhances the softmax loss by introducing an adaptive angular margin penalty between the embedding vectors and their corresponding class centres. 
\textbf{Triplet} loss pulls the anchor closer to the positive image while pushing the negative image away from the anchor. The anchor and positive images share the same class label, while the negative image is derived from an arbitrary class~\cite{hermans2017defense}.
\textbf{\gls{supcon}} loss extends the idea of contrastive learning towards supervised learning~\cite{khosla2020supervised}. The objective is to pull multiple representations of the same class closer together while pushing different classes away in the embedding space. 
Following previous work, we also compare our strategy to \textbf{BioClip} and \textbf{TransReId} a pre-trained classification and re-identification model, respectively~\cite{vcermak2024wildlifedatasets, algasov2024understanding, stevens2024bioclip}. We adopt the \gls{ssl} version for comparison. 

\subsection{Datasets}

The supervised models are pre-trained using the wildlife training set designed by Luk{\'a}{\v{s}} \textit{et al.}~\cite{adam2024wildlifereid}. 
We only use the wildlife data and omit the marine, domestic and livestock animals when training. 
We also use the Wildlife datasets toolkit for training~\cite{vcermak2024wildlifedatasets}.
We pre-train the self-supervised models using camera trap data from two private safaris in South Africa. We detect all the animals in each frame using MegaDetector~\cite{beery2019efficient}. Detections made at a confidence score greater than 0.5 are kept. We collect several subsequent frames at least 120 seconds apart for each frame. Assuming that the individual would have left the scene after two minutes. A shorter interval would increase the number of true positives but reduce the total number of retrieved pairs. Then, we apply \gls{iou} to find any bounding box in the subsequent frames with 0.2 IoU of the bounding box in the starting frame. We consider all candidates that exceed the 0.2 \gls{iou} threshold. Two limitations of this method may arise. First, an image from the previous frame may have been paired with two or more animals that meet the threshold. Second, occlusion or animals entering and exiting the frame can lead to identity drift issues. To mitigate over-engineering solutions to these limitations, we leave it to the model to decipher the correct association. Figure~\ref{figure:pairs} is an illustrative sample of the temporal image pairs obtained using our strategy.

\subsection{Evalution Metrics}

We use a retrieval metric \gls{map} for open-world learning. An \gls{map} of one means all relevant individuals are ranked higher on the list. For the downstream task, we examine the quality of representations by freezing the backbone encoder and obtaining a feature vector for each image~\cite{wu2018unsupervised}. We use a parametric and non-parametric approach~\cite{caron2021emerging, he2020momentum, chen2020simple}.
Weighted \gls{knn} is a non-parametric clustering algorithm~\cite{wu2018unsupervised}. The generated features are stored in a database from which \gls{knn} selects the $k$-nearest features and votes for a relevant label. We fix $k=200$ similar to the work done by~\cite{wu2018unsupervised}.
Linear probing trains a simple linear classifier on top of a frozen backbone encoder. We apply random cropping and horizontal flips during training. After obtaining predicted values, we report on the top-one accuracy. For downstream tasks, we evaluate the model's performance using task-specific metrics. Specifically, \gls{iou} for object detection, Mean IoU for segmentation, Multilabel Accuracy for attribute data, and \gls{pck} for pose estimation.

\subsection{Implementation Details}

The self-supervised models are adopted from lightly, a computer vision \gls{ssl} framework~\cite{susmeljlightly}. Each \gls{ssl} model runs for 100 epochs. We use the same hyperparameters as lightly and do not perform any hyperparameter tuning. We adopt the same augmentation strategies proposed in the original \gls{ssl} papers.
All supervised and self-supervised variants use a Vision Transformer Tiny as the backbone encoder. We only perform hyperparameter tuning on the supervised variants using grid search for 15 epochs and 25 trials. The parameters that produced the lowest loss over the validation set were selected. All training configurations and settings are appended in the supplementary material. We use a single NVIDIA GeForce GTX 1080 GPU to accelerate training.

\begin{table}[htbp]
\caption{Open world re-identification.}
\begin{center}
\resizebox{\columnwidth}{!}{%
\begin{tabular}{lrrr|rrr}
\toprule
& \multicolumn{3}{c}{\makecell{\textit{supervised}\\\textit{(in-distribution)}}} & \multicolumn{3}{c}{\makecell{\textit{self-supervised}\\\textit{(out-of-distribution)}}} \\
\cmidrule(lr){2-4} \cmidrule(lr){5-7}
\textit{\textbf{Species}} & 
    \textbf{Triplet} & \textbf{SupCon} & \textbf{ArcFace} & \textbf{SimCLR} & \textbf{BYOL} & \textbf{DINO}\\
\bottomrule
\toprule
chimpanzee      & 10 & 13 & \cellcolor{blue!20}\textbf{33} & \cellcolor{blue!20}32 & 26 & 26 \\
giraffe         & 13 & 12 & 25 & \cellcolor{blue!20}31 & 29 & \cellcolor{blue!20}\textbf{35} \\
hyena           & 15 & 16 & 31 & \cellcolor{blue!20}35 & 33 & \cellcolor{blue!20}\textbf{37} \\
leopard         & 12 & 21 & \cellcolor{blue!20}\textbf{27} & 20 & \cellcolor{blue!20}22 & 20 \\
nyala           & 07 & 05 & 16 & \cellcolor{blue!20}\textbf{17} & 15 & \cellcolor{blue!20}17 \\
tiger           & 38 & 45 & \cellcolor{blue!20}74 & \cellcolor{blue!20}\textbf{74} & 72 & 74 \\
zebra           & 47 & 49 & 44 & \cellcolor{blue!20}65 & 58 & \cellcolor{blue!20}\textbf{69} \\
Average         & 24 & 27 & 36 & \cellcolor{blue!20}39 & 36 & \cellcolor{blue!20}\textbf{40} \\

\bottomrule
\multicolumn{6}{l}{Top two models are highlighted, with the best model in bold.}
\end{tabular}
}
\label{table:inDist}
\end{center}
\end{table}

\begin{table*}[htbp]
\caption{out-of-distribution open world re-identification.}
\begin{center}
\resizebox{\textwidth}{!}{%
\begin{tabular}{lrr|rrr|rrrrrr} 
\toprule
& \multicolumn{2}{c}{\textit{general-purpose}} & \multicolumn{3}{c}{\textit{supervised}} & \multicolumn{6}{c}{\textit{self-supervised}} \\
\cmidrule(lr){2-3} \cmidrule(lr){4-6} \cmidrule(lr){7-12}
 \textit{\textbf{Models}} & 
 \textbf{TransReID} & \textbf{BioClip} & \textbf{Triplet} & 
 \textbf{SupCon} & \textbf{ArcFace} &
 \textbf{SimCLR} & \textbf{BarlowTwins} & 
 \textbf{MoCo} & \textbf{BYOL} & 
 \textbf{FastSiam} & \textbf{DINO} \\
\midrule
cat             & 50 & \cellcolor{blue!20}\textbf{60} & 20 & 16 & 36 & \cellcolor{blue!20}53 & 45 & 36 & 48 & 41 & 51 \\
cow             & 29 & 30 & 13 & 10 & 22 & \cellcolor{blue!20}\textbf{33} & 28 & 24 & 29 & 26 & \cellcolor{blue!20}31 \\
dog             & 27 & \cellcolor{blue!20}28 & 06 & 04 & 16 & \cellcolor{blue!20}\textbf{28} & 23 & 14 & 22 & 17 & 24 \\
elephant        & 10 & \cellcolor{blue!20}\textbf{15} & 02 & 02 & 07 & \cellcolor{blue!20}13 & 09 & 05 & 09 & 07 & 11 \\
leopard         & \cellcolor{blue!20}17 & \cellcolor{blue!20}\textbf{18} & 08 & 13 & 14 & 14 & 13 & 12 & 13 & 10 & 13 \\
panda           & \cellcolor{blue!20}07 & 06 & 02 & 02 & 04 & \cellcolor{blue!20}\textbf{07} & 06 & 04 & 06 & 05 & 06 \\
polar bear      & 21 & 24 & 13 & 13 & 20 & \cellcolor{blue!20}\textbf{29} & 24 & 19 & 24 & 21 & \cellcolor{blue!20}24 \\
zebra           & 20 & 14 & 04 & 10 & 13 & \cellcolor{blue!20}21 & 17 & 17 & \cellcolor{blue!20}\textbf{21} & 18 & 19 \\
Average         & 23 & \cellcolor{blue!20}24 & 08 & 09 & 16 & \cellcolor{blue!20}\textbf{25} & 20 & 16 & 21 & 18 & 23 \\
\bottomrule
\multicolumn{11}{l}{Top two best-performing models along the column are highlighted. Best model in bold}
\end{tabular}
}
\label{table:outDist}
\end{center}
\end{table*}

\section{Results}\label{section:Results}

\subsection{Open World Re-Identification}

\subsubsection{\textbf{In-Distribution}} 

Table~\ref{table:inDist} presents the in-distribution results for unseen animal classes. The evaluation data was put together by~\cite{adam2024wildlifereid}. The authors allocated a few animal individuals in the test set but not in the training set, such that they are derived from the same distribution but have not been seen by the supervised models. All the classes are out-of-distribution for the self-supervised models because they were trained using camera trap data. We present the results for contrastive learning, self-distillation, and the current SOTA strategy. The remaining methods yield similar results, primarily determined by the strategies they employ.
The results demonstrate that self-supervised methods consistently outperform all supervised variants. \gls{simclr} performs nearly on par with \gls{arcface} on the tiger's dataset (74.41 vs 73.98 mAP) despite the absence of tigers in the camera trap training data. Self-supervised learning maintains its superior performance even under data-limited conditions.

\subsubsection{\textbf{Out-Of-Distribution}} 

Table~\ref{table:outDist} reflects on the out-of-distribution results. The datasets are new and have not been used for training. 
\gls{simclr} achieves the highest \gls{map}, with BioClip and TransRec following closely behind. The general-purpose model's performance is bolstered by its extensive parameter count. BioClip and TransRec are approximately 27 and 4 times larger than ViT-Tiny. Notably, the self-supervised variants outperform supervised models designed for wildlife re-identification tasks. Triplet and \gls{supcon} loss are roughly half as good as \gls{arcface} loss. Their use of negative labels may result in overfitting over the seen classes. However, \gls{simclr} performs optimally with the use of negatives. \gls{dino} ranks as the third best-performing model and demonstrates the most consistent performance. 

\begin{table*}[htbp]
\caption{Downstream Tasks and Out-of-Domain Results}
\begin{center}
\resizebox{\textwidth}{!}{%
\begin{tabular}{lrrrrr|rrr|rr|rr}
\toprule
& \multicolumn{4}{c}{\textbf{image}} & \multicolumn{1}{c}{\textbf{video}}
& \multicolumn{1}{c}{\textbf{Detection}} & \multicolumn{2}{c}{\textbf{Segmentation}}
& \multicolumn{1}{c}{\textbf{Attribute}} & \multicolumn{1}{c}{\textbf{Pose}} & \multicolumn{2}{c}{\textbf{Out-of-Domain}} \\
 \textbf{Method} &
 \multicolumn{2}{c}{\textit{iNat21}}~\cite{van2021benchmarking} & 
 \multicolumn{2}{c}{\textit{CIFAR}}~\cite{krizhevsky2009learning} & 
 \textit{AK}~\cite{ng2022animal}
 & \multicolumn{1}{c}{\textit{Animals}}~\cite{jana2023detection} & \textit{Coco}~\cite{lin2014microsoft} & \textit{Oxford Pets}~\cite{parkhi12a} & \multicolumn{1}{c}{\textit{AWA}}~\cite{xian2018zero} & \textit{AK}~\cite{ng2022animal} & \textit{Plants}~\cite{nilsback2008automated} & \textit{Insects}~\cite{van2021benchmarking} \\
 \cmidrule(lr){2-3} \cmidrule(lr){4-5} \cmidrule(lr){6-6} \cmidrule(lr){7-7} 
 \cmidrule(lr){8-8} \cmidrule(lr){9-9} \cmidrule(lr){10-10} \cmidrule(lr){11-11} \cmidrule(lr){12-12} \cmidrule(lr){13-13}
& KNN & Linear & KNN & Linear & \multicolumn{1}{c}{mAP} 
& \multicolumn{1}{c}{IoU} & \multicolumn{2}{c}{mIoU} & \multicolumn{1}{c}{Multi-Label} & \multicolumn{1}{c}{PCK} & \multicolumn{2}{c}{KNN} \\
 \toprule
\textbf{\textit{general purpose}} \\
TransReID           & \cellcolor{blue!20}11.2 & \cellcolor{blue!20}16.2 & 36.4 & 40.0 & 47.2 & \cellcolor{blue!20}48.7 & 2.80  & 43.8 & \cellcolor{blue!20}71.3     & \cellcolor{blue!20}\textbf{47.8} & \cellcolor{blue!20}44.5 & \cellcolor{blue!20}1.6\\
BioClip             & \cellcolor{blue!20}\textbf{60.0} & \cellcolor{blue!20}\textbf{68.8} & \cellcolor{blue!20}\textbf{41.5} & \cellcolor{blue!20}\textbf{89.4} & \cellcolor{blue!20}47.6 & \cellcolor{blue!20}\textbf{55.4}    & \cellcolor{blue!20}\textbf{3.00} & 36.9 & \cellcolor{blue!20}\textbf{77.3}     & 32.9  &  \cellcolor{blue!20}\textbf{94.4} & \cellcolor{blue!20}\textbf{60.4}\\
\textbf{\textit{supervised}} \\
Triplet             & 3.20 & 4.50 & 18.2 & 24.2 & 42.0 & 21.2     & 2.80 & 38.0 & 68.1 & 26.7 & 7.30 & 0.20 \\
SupCon              & 3.30 & 3.40 & 19.1 & 18.7 & 39.4 & 21.6     & 2.00 & 38.6 & 67.4 & 23.8 & 12.9 & 0.20 \\
ArcFace             & 4.70 & 5.90 & 26.1 & 30.6 & 43.8 & 30.0     & 2.70 & 39.4 & 68.9 & 29.6 & 18.0 & 0.50 \\
\bottomrule
\textbf{\textit{self-supervised}} \\
SimCLR              & 7.30 & 9.60 & 31.4 & 39.1 & 42.9 & 41.1      & \cellcolor{blue!20}2.90 & 40.3 & 68.8 & 26.5 & 18.9 & 1.20 \\
BarlowTwins         & 5.50 & 6.20 & 28.8 & 31.3 & 41.3 & 26.7      & 2.60 & 39.0 & 64.9 & 26.0 & 8.40 & 0.40 \\
MoCo                & 4.90 & 6.50 & 26.2 & 33.4 & 45.4 & 27.1      & 2.80 & \cellcolor{blue!20}\textbf{46.0}  & 69.4 & 28.8 & 19.3 & 0.50 \\
FastSiam            & 6.30 & 9.40 & 19.5 & \cellcolor{blue!20}40.8 & 44.6 & 32.9      & 2.80 & 41.2 & 69.3 & \cellcolor{blue!20}33.4 & 23.0 & 0.70 \\
BYOL                & 7.60 & 8.90 & 32.9 & 39.1 & 47.2 & 38.9      & 2.90 & 41.2 & 69.8 & 32.3 & 31.0 & 1.00 \\
DINO                & 7.90 & 9.70 & \cellcolor{blue!20}36.4 & 40.5 & \cellcolor{blue!20}\textbf{48.2}  & 38.2      & 2.70 & 41.7 & 69.6 & 33.2 & 28.6 & 0.90 \\
\bottomrule
\multicolumn{6}{l}{Top two along the columns are highlighted. Best model in bold}
\end{tabular}
}
\label{table:combined}
\end{center}
\end{table*}

\subsection{Downstream Tasks}

Table~\ref{table:combined} presents the downstream performance of general-purpose, supervised and self-supervised learning methods. We freeze the encoders pre-trained for the re-identification task. These representations are consistent across non-overlapping frames at different locations. This strategy ensures that classification, detection, segmentation and pose-estimation are applied to the same instance.

\subsubsection{\textbf{Image Classification}} 

\gls{dino} once again demonstrates the strongest performance among the self-supervised variants. \gls{arcface} remains the top-performing supervised model. However, it underperforms by four per cent and two per cent compared to the weakest self-supervised model, Barlow Twins, on iNat21 and CIFAR, respectively. Notably, ReID models achieve only a fraction of the performance of BioClip. This gap arises because ReID models rely heavily on local-level features, which limits their ability to learn global, category-level representations.

\subsubsection{\textbf{Video Classification}} 

We extract multiple frames from the \gls{ak} videos dataset, pass them through a backbone encoder, and average the resulting features for linear probing. 
\gls{dino} achieves a $2\%$ higher \gls{map} than BioClip, establishing it as the top-performing model. \gls{byol} ranks third, trailing BioClip by only 0.82\%. These results demonstrate that self-supervised models can effectively track individuals over time, a capability we attribute to our temporal image-pair training strategy. Interestingly, the best-performing models (\gls{dino}, \gls{byol}, and \gls{moco}) all employ dual networks without shared weights.

\subsubsection{\textbf{Object Detection}}

Object detection requires features that are scale and rotation-invariant. Each feature should also contain some object-specific information so that the bounding box is later classified. 
Self-supervised models achieve an IoU score that is almost double that achieved by Triplet and \gls{supcon} loss. \gls{simclr}, \gls{byol} and \gls{dino} show the strongest performance.

\subsubsection{\textbf{Image Segmentation}}

We use the \gls{miou} to measure the average overlap between the predicted segmentation masks and the ground truth label across all classes. \gls{moco} performs strongest for the semantic segmentation task on OxfordPets and comparatively on CoCo~\cite{parkhi12a, lin2014microsoft}. Semantic segmentation requires stronger pixel-level features. The features should have a good global and local contextual understanding of the image. The global features capture the overall scenery, and the local features encode the edges and textures. \gls{dino} achieves second place, followed by the self-supervised methods on OxfordPets, but struggles with Coco. OxfordPets has three classes, while Coco has 12. All models struggle to cluster cluttered images.

\subsubsection{\textbf{Attribute Data}}

Attribute data describes an animal's characteristics in tabular format, representing a multi-label learning task. Since a single data point can be mapped to multiple classes, we use the multi-label accuracy. This metric computes the average top-one accuracy over each attribute. 
Supervised and self-supervised models perform comparably, with \gls{dino} achieving only $1\%$ higher accuracy than \gls{arcface}. Moreover, self-supervised features remain as descriptive as supervised features, even without labels.

\subsubsection{\textbf{Pose Estimation}}

Pose estimation involves predicting the spatial positions of key body joints. \gls{pck} measures the model's ability to predict keypoint locations accurately. Specifically, it computes the average number of predicted keypoints that sit within some distance from the ground truth. Among the evaluated methods, TransReID achieves the highest \gls{pck} score of 47.8, outperforming \gls{fastsiam} and \gls{dino}, which scored 33.4 and 33.2, respectively. The superior performance of TransReID in pose estimation highlights the advantages of leveraging large-scale models trained on extensive datasets. 

\subsubsection{\textbf{Out-of-Domain}}

\gls{byol} and \gls{dino} outperform the best-supervised model by $72\%$ and $58\%$ on the Flowers dataset, respectively. Flowers typically share similar shapes but exhibit distinct colours and patterns.
However, performance on the insect's dataset is relatively marginal for both self-supervised and supervised models, with the top-performing models (\gls{byol} and \gls{dino}) achieving scores of only 0.0104 and 0.0094. The small size of insects likely necessitates a global representation rather than fine-grained local features. 

\subsection{Ablation}

\subsubsection{\textbf{Temporal image Pairs}} 

We conduct an ablation study to evaluate the impact of dataset size on model performance. \gls{simclr} is chosen for this evaluation due to its lightweight architecture and consistent performance. Figure~\ref{figure:ablation} illustrates the \gls{map} achieved by \gls{simclr} across varying \gls{iou} threshold values. 
A lower threshold value yields a larger number of pairs. The results describe a general downward trend in performance as the threshold increases. The lowest mAP is achieved at a threshold of 0.2.

\subsubsection{\textbf{Temporal Image Pairs and Self Distillation}} 

We further expand the training dataset by integrating our strategy with self-distillation. Self-distillation is the proposed training approach for \gls{simclr}~\cite{chen2020simple}. They generate two views by applying augmentations to a single image. The combined temporal pairing and self-distillation strategy outperforms temporal pairing alone across all threshold values. However, the performance trend remains relatively stable as the IoU threshold increases.
We only conduct the ablation study on \gls{simclr} due to its fast training process. However, in our experiments, we observed a similar trend \gls{moco} and BarlowTwins. 

\begin{figure}[htp]
  \centering
  \includegraphics[width=1.0\linewidth]{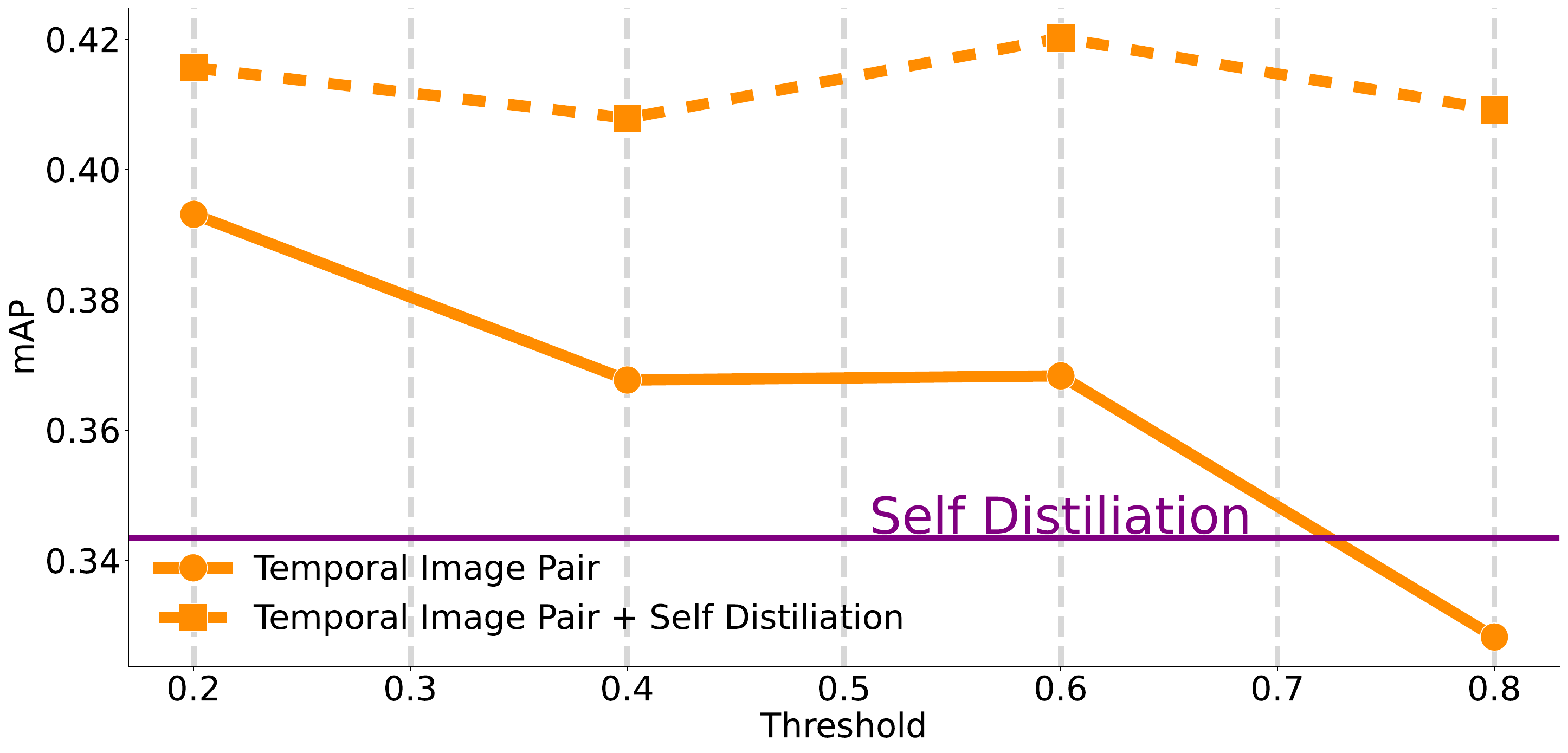} 
  \caption{\gls{map} achieved by \gls{simclr} across increasing \gls{iou} thresholds. Lower thresholds produce more temporal pairs, enhancing model performance. Combining temporal pairing with self-distillation further expands the training dataset, achieving optimal results.}\label{figure:ablation}
\end{figure}

\begin{figure}[htp]
  \centering
  \begin{tabular}{ccccc}
  \includegraphics[width=0.4\linewidth]{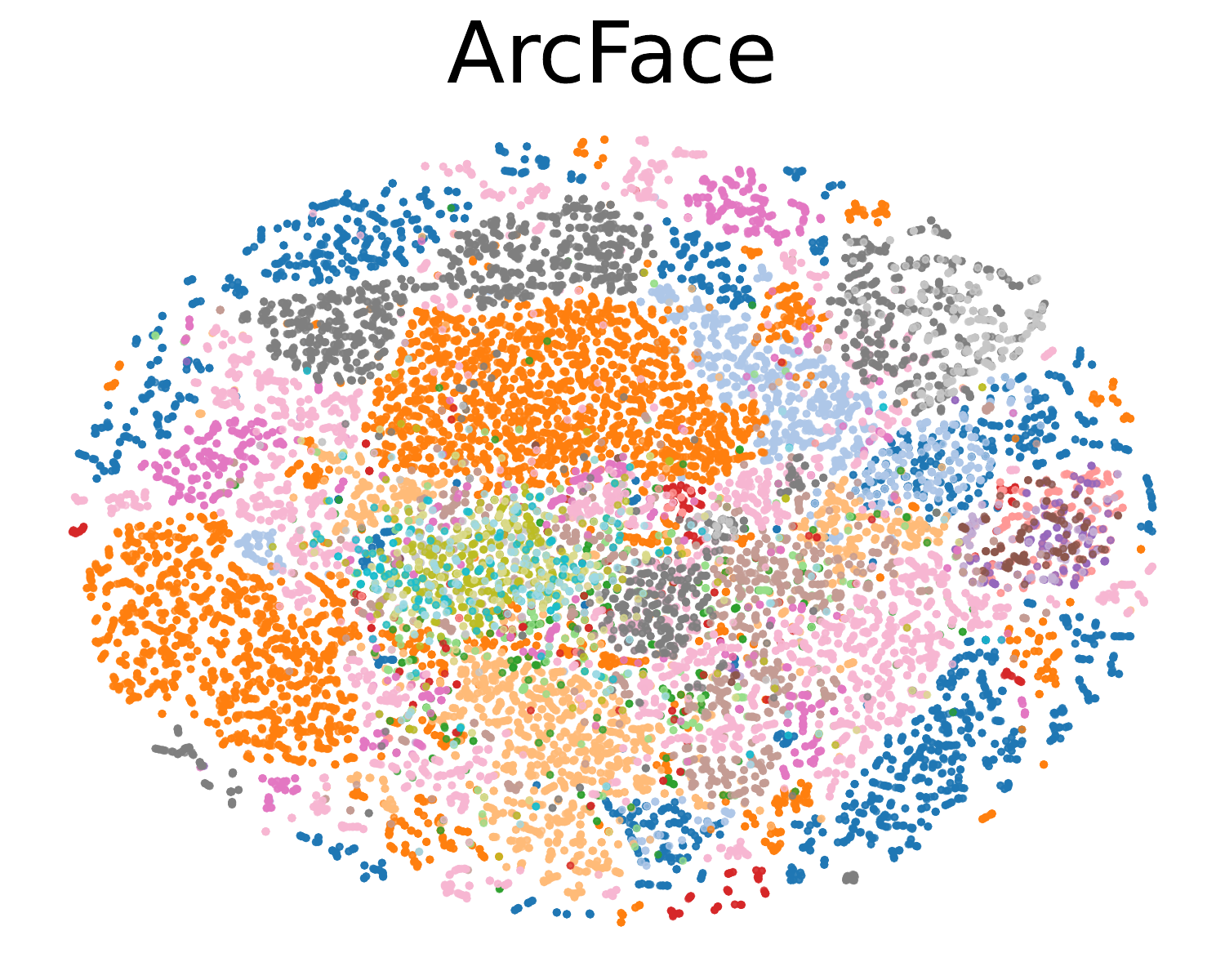} &
  \includegraphics[width=0.4\linewidth]{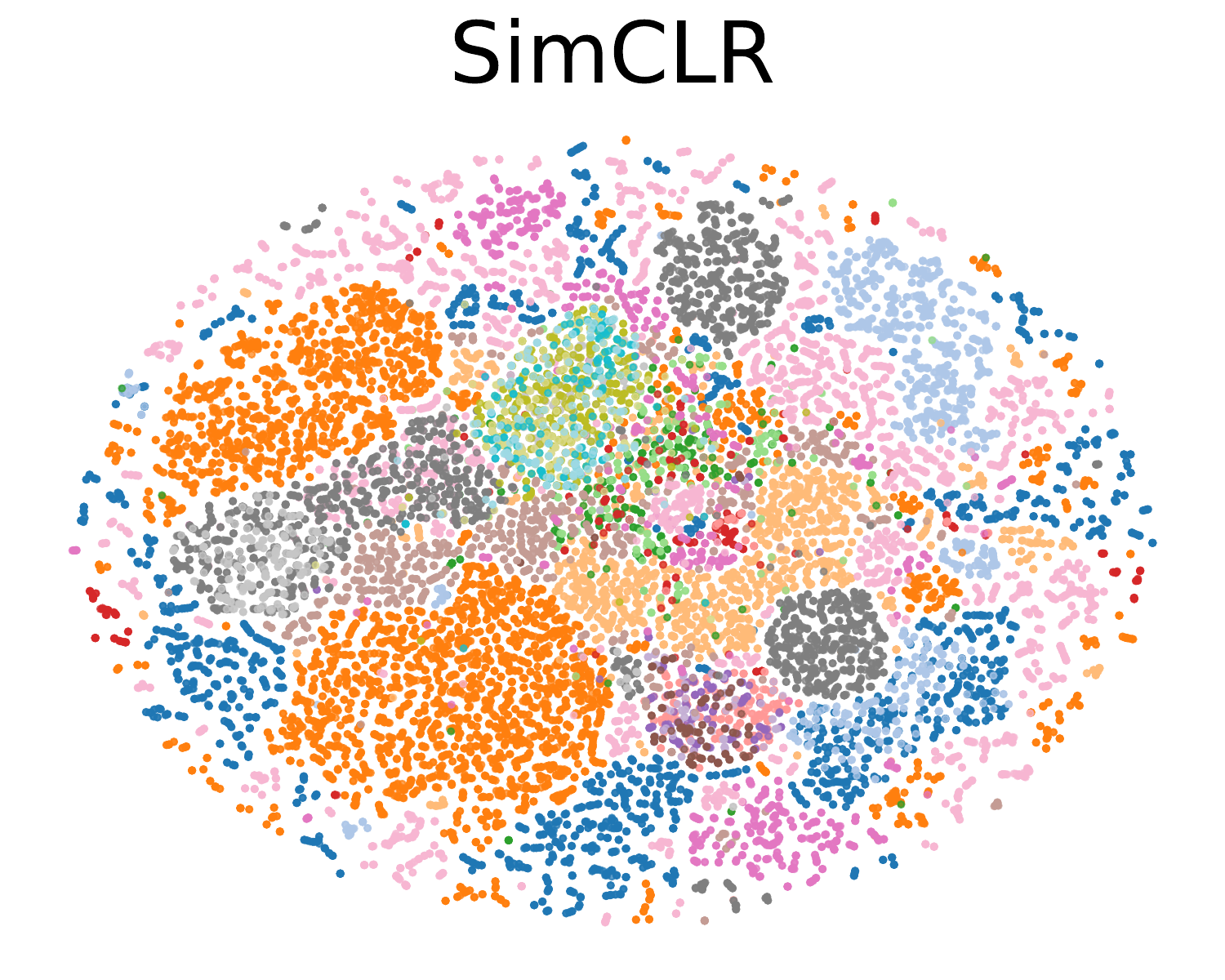} \\
  \end{tabular} 
  \caption{Latent space separability of \gls{arcface} and \gls{simclr} on open-world data. \gls{simclr} forms more distinct clusters with minimal overlap. All \gls{ssl} models demonstrate comparable visual coherence in their latent representations.}\label{figure:latent}
\end{figure}

\begin{figure}[htp]
  \centering
  \includegraphics[trim={0 11.88cm 0 0},clip,width=1.0\linewidth]{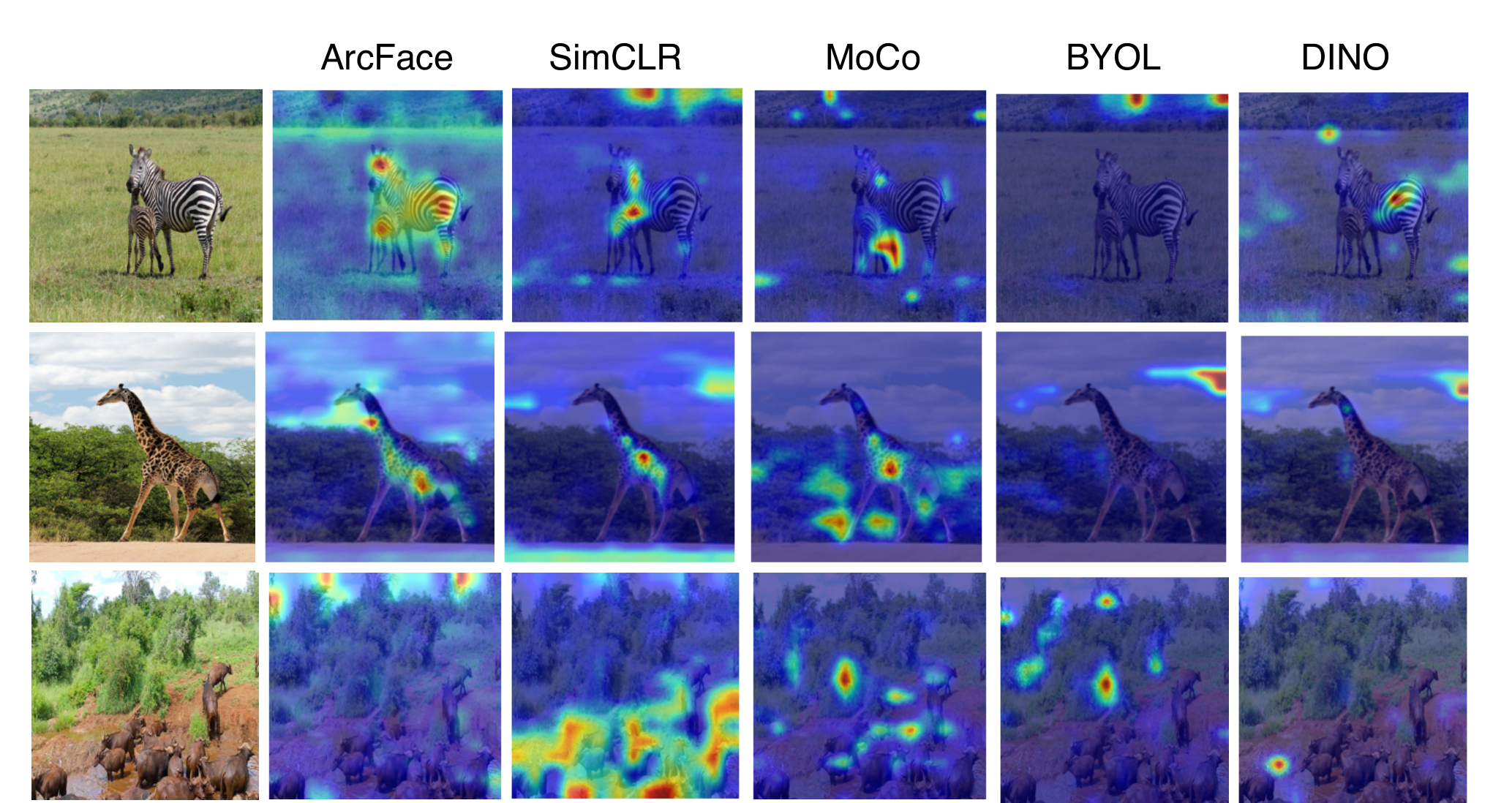}
  \caption{Attention maps of models on images within natural environments. \gls{simclr} consistently identifies all individuals, followed by \gls{moco}. \gls{byol} and \gls{dino} perform best at close range but rely on background at greater distances. \gls{arcface} struggles with cluttered scenes.}\label{figure:attention}
\end{figure}

\subsection{Qualitative Analysis}

\subsubsection{\textbf{Latent Space}}

Figure~\ref{figure:latent} examines the latent space separability of each model. We use the open-world data designed by~\cite{adam2024wildlifereid}. Self-supervised models exhibit more distinct clusters with minimal overlap between points. The latent space demonstrates strong separability despite the absence of labelled training data. 

\subsubsection{\textbf{Attention Map}}

Figure~\ref{figure:attention} illustrates the attention maps generated by each model for images without bounding boxes. 
\gls{arcface} loss struggles to locate an individual when multiple animals, objects or people are within the image. Negative-based methods like \gls{simclr} and \gls{moco} show the strongest ability to find animals in the image. \gls{byol} and \gls{dino} perform best when animals are closer to the camera, but their reliance on background scenery increases with greater distances. 

\section{Conclusion}\label{section:Conclusion}

The study proposed a simple but effective strategy to extract temporal pairs for self-supervised wildlife re-identification automatically.
The results show that self-supervision generates robust features under wildlife open-world scenarios and generalizes well over several downstream tasks. 
Future research should study extracting multiple views of an individual from camera traps with video-based self-supervised learning. 
Continual learning could also enhance the quality of features by enrolling new classes into pre-trained models~\cite{huo2025incremental}

\subsection{Broader Impact}

The study proposes a method to automatically extract temporal image pairs from camera traps for self-supervised learning. Researchers can now develop robust representations effective in shifting distributions, multiple downstream tasks, and out-of-domain tasks.

\subsection{Limitations}

This study only evaluates each model in the open world and transfers learning datasets. We do not consider the in-distribution scenario as this has been substantially covered in previous work and has shown considerable success. Although \gls{ssl} demonstrates sustained performance improvements over supervised models, it still struggles with fine-grained feature discrimination. This is evident in the marginal performance for leopards and insects. Furthermore, the giraffe attention map in Figure~\ref{figure:attention} reveals that \gls{ssl} rely on background cues as the camera distance increases. Future work could explore finer-resolution encoders or video-based \gls{ssl} strategies. 

\section*{Acknowledgment}

We extend our gratitude to Qulinda AB and Linköping University, Sweden, for providing the camera trap data used for training. This work was funded by the \gls{dti}, South Africa, with additional support from the \gls{cair}.

\bibliographystyle{IEEEtran}
\bibliography{main}

\end{document}